\documentclass{article}

\usepackage{PRIMEarxiv}

\usepackage{cite}
\usepackage{amsmath,amssymb,amsfonts}
\usepackage{algorithm}
\usepackage[noend]{algorithmic}
\usepackage{graphicx}
\usepackage{textcomp}
\usepackage{booktabs}
\usepackage{epstopdf}
\usepackage{fancyhdr}  

\title{Comparative assessment of federated and centralized machine learning}
	
\author{
	Ibrahim Abdul Majeed \\
	Samsung R\&D Institute India – Bangalore  \\
	\And
	Sagar Kaushik \\
	Birla Institute of Technology and Science,  \\
	Pilani, India\\
	\AND
	Aniruddha Bardhan \\
	Indian Statistical Institute, Kolkata, India\\
	\And
	Venkata Siva Kumar Tadi \\
	Samsung R\&D Institute India – Bangalore  \\
	\And
	Hwang-Ki Min \\
	Samsung Electronics, Suwon-si, Korea \\
	\And
	Karthikeyan Kumaraguru \\
	Samsung R\&D Institute India – Bangalore \\
	\And
	Rajasekhara Duvvuru Muni \\ 
	Samsung R\&D Institute India – Bangalore \\
}

\begin{document}
		\maketitle
	\def\thefootnote{*}\footnotetext{Sagar Kaushik and Aniruddha Bardhan contributed to the work during their internship at Samsung Research and Development Institute, Bangalore }\def\thefootnote{\arabic{footnote}}
	
%
	
	
	\begin{abstract}
		Federated Learning (FL) is a privacy preserving machine learning scheme, where training happens with data federated across devices and not leaving them to sustain user privacy. This is ensured by making the untrained or partially trained models to reach directly the individual devices and getting locally trained ``on-device" using the device owned data, and the server aggregating all the partially trained model learnings to update a global model. Although almost all the model learning schemes in the federated learning setup use gradient descent, there are certain characteristic differences brought about by the non-IID nature of the data availability, that affects the training in comparison to the centralized schemes.  In this paper, we discuss the various factors that affect the federated learning training, because of the non-IID distributed nature of the data, as well as the inherent differences in the federating learning approach as against the typical centralized gradient descent techniques. We empirically demonstrate the effect of number of samples per device and the distribution of output labels on federated learning.
		In addition to the privacy advantage we seek through federated learning, we also study
		if there is a cost advantage while using federated learning frameworks. We show that federated learning does
		have an advantage in cost when the model sizes to be trained are not reasonably large. All in all, we present the need for careful design of model for both performance and cost.

	\end{abstract}
	
	
	
	\section{Introduction}
	\label{sec:introduction}
	 Hand held gadgets like mobile phones and tablets have become part of every day life for most of the world population. Over the course of time, these small devices have also become a great source of data, as well as computationally robust units.
	 Together with the recent abundance of data, both quantitatively and quantitatively, data privacy has become one of the major area of concern. Users would love to have more engaging and personalized experiences with the gadgets, but at the same time will be privy in sharing their data for using it to build better applications. With more and more stricter rules like the General Data
	Protection Regulation (GDPR)~\cite{regulation2016regulation}, the usage of private data need to be handled with utmost care ensuring minimal privacy leakage. At the same time, the main part of developing any learning algorithm is the existence of data, from which
	salient features pertaining to the use-case can be learned. 
	Federated learning (FL)~\cite{mcmahan2017communication, yang2019federated, bonawitz2019towards} is a recent area of research, where the learning happens with the  private data  not leaving the
	device. 
	
	In this setting, a shared global model under the supervision of a central orchestrator or server (called aggregator)\footnote{Since the aggregator resides in the server, in the federated learning scenario, we will call it as server, aggregator or central aggregator.} gets trained
	by a federation of participating devices.
	In a typical FL training environment, a base model is initiated by a server, which is then pushed to the
	participating devices. The devices updates the model parameters, individually, in a typical optimization process like
	stochastic gradient descent (SGD) or mini-batch gradient descent, involving batches and epochs, then returns the
	parameter weight updates back to the aggregator. The aggregator then aggregates all the received weight updates and updates the
	parameters of the base model. The process is repeated until the model converges satisfactorily with the desired performance criterion.

	The FL process briefly explained above is fundamentally not the same as the original gradient descent algorithm design. However special
	cases of the schemes do match some of the gradient descent variants.  Still, there are certain challenges that we face with 	respect to the learning of models, because of the non IID nature of the data. The typical federated nature of data may introduce new learning challenges like each device having single labeled data or even single sample. In addition, various other factors like  number of samples per device,  the feature space, the model complexity all have an effect on performance of models trained via federated learning. 
	
	While considering the infrastructure cost involved in federated learning training, we see that the main advantage of FL is because of its primary security design, that is we need not have to 	sync the data to a central server\cite{bonawitz2019towards}. Syncing data to a central server involves a cost, which may not be negligible. In the case of FL training, the major chunk of the cost involved are related to sending/receiving the model parameter updates to/from the server
	aggregator: the more complex the model, the more the number of parameters related to the model that need  to be
	send by the devices to the aggregator, and hence more the corresponding model update data that needs to be synced to the
	central aggregator. More than the cost involved during training is the model deployment cost, which involves not just
	the devices participating in the training, but all the devices to which the inference engines has to be deployed. Thus,
	keeping the model size under control is essential for reducing the cost in FL.
	
	The another aspect of FL training is security. Although the data is ensured not to leave the devices, we also have to
	make sure that the updates are also secure, so as to reduce the chances of inference of data from gradients~\cite{geiping2020inverting}. Typical
	private AI schemes like differential privacy~\cite{dwork2011firm, dwork2014algorithmic,dwork2006calibrating, baek2021enhancing}, secure aggregation~\cite{bonawitz2017practical} and homomorphic encryption~\cite{hardy2017private} have been employed in FL
	in varying degrees for secure communication  of the training updates. However, adding security to the process affects the model
	performance or the training duration, thus impacting the cost. As we try to anonymise the data, the more the information loss is and hence lesser or slower would be the learning~\cite{mcmahan2017learning}.

	In this work, we address two main problems:
	\begin{enumerate}
		\item How does federated learning optimization algorithms work in comparison to traditional gradient descent based
		optimization techniques for specific non-IID settings, involving skewness in data samples and labels.
		\item How much cost effect does federated learning based training and deployment have on the cloud platforms as
		compared to centralized learning and inference schemes.
	\end{enumerate}
	
	Recent works have looked at how federated learning performs in general, for different non-IID scenarios. We try to extend their work by considering more practical scenarios related to on-device data. We first show theoretically, how FL is different from central learnings and how the federated averaging algorithm~\cite{mcmahan2017communication} is technically different from gradient descent variants.
	We look at some of the special non-IID cases associated with number of samples and skewness of labels and see how it affects learning. We also show a comparison of costs involved in federated learning under different scenarios and centralized\footnote{The term ``central'', when used in the federated learning context would mean the central aggregator or server. General usage of ``central'' or ``centralized'' would mean the training happening with all the data available at one place.} schemes. 
	
	The rest of the manuscript is arranged as follows. Section~\ref{sec:related} discusses the  related works in the area of federated learning. In section~\ref{sec:FL}, we discuss how federated learning is practically different from centralized training. The various experiments supporting section section~\ref{sec:FL} is shown in section~\ref{sec:exp}. We show the cost comparison between federated and centalised schemes in section~\ref{sec:cost}, followed by conclusion.
	
	\vspace{5pt}
	\section{Related works}
	\label{sec:related}
	Federated learning as a concept was introduced by McMahan {\it et al.}~\cite{mcmahan2017communication} in $2016$, although privacy preserving computations were already discussed in the 80's~\cite{rivest1978data, yao1982protocols}.  
	There  have also been earlier works on privacy preserving techniques using local data with the help of central servers~\cite{agrawal2000privacy, vaidya2008privacy}.
	
	
	Initially coined for edge devices or mobiles, federated learning has also been extended to other domains where edge clients would be large organizations and would be relatively small in number. The setting involving federation of many edge devices with relatively few data samples per device is called cross-device federated learning. The second setting involving few number of large organisations that themselves act as data centers is called a cross-silo federated learning. An example of cross-silo is medical image classification using hospital data~\cite{kaissis2020secure}. Here each hospital acting as a client has large non-sharable data, yet would like to learn a medical image classification model in collaboration with other hospitals. In this work, we would not analyze the learning challenges and cost effects of cross-silo setting. The survey paper by Kairouz {\it et al.} ~\cite{kairouz2019advances}, discusses the different federated settings, open problems and challenges in detail.
	
	Since the conceptualization of federated learning, it has been widely used in many domains. As a primary initiator, Google has incorporated federated learning in its GBoard keyboard~\cite{hard2018federated, ramaswamy2019federated, yang2018applied}, in android messaging and in web tracking as a replacement for third party cookies~\cite{ravichandran2021evaluation, bindra2021building}. Cross-silo usecases include medical data segmentation and classification~\cite{sheller2020federated, kaissis2020secure}, financial risk prediction, pharmaceutical discovery, etc.
	
	In terms of analysis of federated learning algorithm, the original authors~\cite{mcmahan2017communication} showed FL performance in various scenarios including IID and non-IID distribution of data, with increasing the number of rounds, and increasing the number of clients. In almost all the experiments, their federated averaging method was shown to be better than the naive federated SGD algorithm. \cite{stich2018local, yu2019parallel}  look at asymptotic local SGD convergence and \cite{kairouz2019advances} extends it for the federated setting in their survey.
	
	Different federated learning frameworks have also come up in the recent years with the important ones being  TensorFlow Federated~\cite{The_TensorFlow_Federated_Authors_TensorFlow_Federated_2019}, 	PySyft~\cite{ryffel2018generic}, Federated AI Technology Enabler~\cite{The_FATE_Authors_2019}, Leaf~\cite{caldas2019leaf}. There have also been attempts to benchmark a few of these  frameworks also~\cite{The_TensorFlow_Federated_Authors_TensorFlow_Federated_2019, budrionis2021benchmarking}.
	
	In our work, we look into the training aspects of federated learning, when there is label sparsity; i.e., when we have only single label available per edge device for a multi-class learning problem.  We also give a detailed study of the cost factor associated with federated learning, which is not a well covered issue. We hope that this will help FL designers to come up with cost optimal federated solutions or even to decide whether federated learning is actually required.

%
	
	\vspace{5pt}
	\section{Optimization in Federated `Learning'}
	\label{sec:FL}
	Federated learning optimization schemes use gradient updates for improving the model during training. The gradient updates happen both at the server end (global), as well as at the client end (local). The basic global
	model gradient update step in federated learning is similar to any gradient update step, denoted as
	\begin{equation}
		\label{eqn:weight_update}
		w^{t+1} = w^t - \eta_s \cdot f(gradient)
	\end{equation} 
	What differs is how we compute the gradient  function $f(gradient)$, which is basically a function of the client
	gradients computed for each round of FL training.
	In this section, we see the optimization schemes usually employed in federated learning. We see how it is
	different from the normal gradient descent based optimization schemes. This will help in understanding the
	practical difficulties in federated learning in comparison to centralized schemes.
	But first, we will briefly overview the traditional gradient descent schemes, which will help in mapping the federated learning
	gradient
	descent based optimization schemes more clearly.
	
	\subsection{Traditional gradient descent learning}
	Any learning algorithm requires an objective function to optimize. In case of most of the classification and regression
	problems, the objective function is the loss function, which is minimized. This loss function is actually an
	aggregation
	of the individual losses associated with each sample used for training. Let's assume the training data set to be
	$\mathcal{D} = \{(x_i, y_i)\}_{i=1}^{N}$, where $N$ is the total number of training samples, $x_i, y_i$ are the input representation and output label, respectively of the $i$th samples. Similar to the notation in \cite{mcmahan2017communication}, we denote the optimization problem to be
	
	\begin{equation}
		\label{eqn:obj}
		\min_{w \in \Re^d} f(w)\mbox{ where } f(w) = \frac{1}{N} \sum_{i=1}^{N} {l}(x_i, y_i, w)
	\end{equation}
	where $l(x_i, y_i, w)$ is the loss associated with input-output pair ($x_i, y_i$) with model parameter $w$.	A gradient
	step involves calculation of gradient of $f(w)$, denoted as 
	\begin{equation}
		\label{eqn:grad}
		g(w) = \nabla f(w) = \frac{1}{N}\sum_{i} g_i(w)
	\end{equation} where 
	\[g_i(w) = \frac{\partial {l}(x_i, y_i, w)}{\partial w}\]

	Note that while optimizing the above loss function, we may not take the entire dataset for each iteration. When only a subset of samples, whose indices denoted by  $B \subset \{1,\ldots N\}$ is used, the loss function would be
	\begin{equation}
		\label{eqn:obj_batch}
		\min_{w \in \Re^d} f_B(w)\mbox{ where } f_B(w) = \frac{1}{n_B} \sum_{i \in B} {l}(x_i, y_i, w)
	\end{equation}
	where $n_B = |B|$.
	The gradient associated with this loss function is \[g_B(w) = \nabla f_B(w) = \frac{1}{n_B}\sum_{i \in B} g_i(w)\]
	
	In gradient descent based optimization schemes, each step of weight updates correspond to a gradient computation step involving samples in a batch. The model learning  happens over weight updates for batches of training samples. When all samples (in batches) are presented for training once, it corresponds to one epoch. The number of epochs is an indication of the amount of training 
	\begin{itemize}
		\item  When all the samples in $\mathcal{D}$ are considered  as a batch, it is called batch gradient
		descent. This would correspond to the case when $n_B = N$ given in ~\eqref{eqn:obj} and ~\eqref{eqn:grad}.
		\item When a randomly picked batch of samples, subscripted by $B$ is considered, it is called stochastic gradient descent. 
		\item When we consider only a single sample per step, which may not be stored for later, then it is called online
		gradient descent. This would correspond to the scenario when $N \sim \infty$, and $|B| = 1$.
		
	\end{itemize}
	It is easy to show that when the samples are IID and when the samples in each batch are also IID and drawn from the same distribution as the training data, then $E[f_B(w)] = f(w)$ and $E[g_B(w) = g(w)]$. However, in general, depending on the samples in the batch, which could have a different distribution as the training data, the loss function that is minimized for a batch would be different from the loss function of the entire data. Hence, a single update for a batch $B$ would reduce the loss $f_B(w)$ for that batch, yet increase the error on the full dataset,  $f(w)$. However over a course of different batches, the loss function $f(w)$ in \eqref{eqn:obj} would reduce~\cite{duda2006pattern}.
	
	\subsection{Federated Averaging}
	In centralized setting, each batch is processed sequentially to update the model. In the federated setting,  the data is distributed over clients. As mentioned earlier, two types of gradient updates happen in federated optmization:  one done locally at the client side, involving the data present at clients and one at the server side, which involves aggregation of individual client update to update the global model weights. Depending on how the two gradient updates work at clients and server, the optimization in a federated setting may match the centralized setting or could be quite different. However the final goal would still be the same; to minimize the loss of the whole training data.
	
	Let us assume there are $K$ clients and the $N$ samples are distributed over these clients with $P_k$ the set of indexes of data points on client $k$.
	Thus, we can re-write the objective function \eqref{eqn:obj} as 
	\begin{equation}
		\label{eqn:obj_fed}
		f(w) = \sum_{i=1}^{K}\frac{n_k}{N} f_k(w)\mbox{ where }  f_k(w) = \frac{1}{n_k} \sum_{i \in P_k} {l}(x_i, y_i, w)
	\end{equation}
	where $n_k = |P_k|$
	 
	Let us assume a fraction $C$ of clients is chosen in a particular round of federated learning. When we compute one step of gradient descent at each client using all the data present within them, we call this federated algorithm as \texttt{FederatedSGD}~\cite{mcmahan2017communication}. When $C = 1$, we obtain \eqref{eqn:obj} corresponding to batch gradient descent. 
	
	Applying gradient to ~\eqref{eqn:obj_fed} with $C=1$ gives
	\[
		\nabla f(w) = \sum_{i=1}^{K}\frac{n_k}{N} \nabla f_k(w)
	\]
	This implies the batch gradient is equal to the weighted average of the individual client gradients in \texttt{FederatedSGD} (with $C=1$). Hence using an appropriate step size of $\eta$, the weight update step in \texttt{FederatedSGD}, which would be
	\begin{equation}
		\label{eqn:Fed_wt_update}
		w_{t+1} = w_t - \eta \cdot \sum_{i=1}^{K}\frac{n_k}{N} \nabla f_k(w_t)
	\end{equation}
	will reduce the global loss function.
	
	Similar to the central setting, we could assume that each client passes over its data in batches, $B$, and passes over the entire data, $E$ (epoch) times. This tweak to the \texttt{FederatedSGD} is called the \texttt{FederatedAveraging}~\cite{mcmahan2017communication}. The pseudo-code for \texttt{FederatedAveraging} is provided in Algorithms~\ref{alg:FedAvg} and~\ref{alg:ClientUpdate}. Again, when $\mathcal{B} = \{P_k\}$ for each client, $k$, we get \texttt{FederatedSGD}. Thus \texttt{FederatedSGD} is a special case of \texttt{FederatedAveraging}. The $ClientUpdate()$ function given in Algorithm~\ref{alg:ClientUpdate} is the one which actually drives the learning and rate of \texttt{FederatedAveraging}, as it directly involves the client side data, as well as the skewness in the data, together with the batch size and epochs.

	\begin{algorithm}
		\caption{\texttt{FederatedAveraging} as in~\cite{mcmahan2017communication}. \newline The $K$ clients are
			indexed by $k$}
		
		\begin{algorithmic}
			\STATE initialize $w_0$
			\FOR {each round $t = 1,2, \ldots$ }
				\STATE $m \gets max(C \cdot K, 1)$
				\STATE $S_t\gets (\mbox{random set of m clients})$
				\FOR {each client $k \in S_t$ \textbf{in parallel}} 
					\STATE $w_{t+1}^k \gets  ClientUpdate(k, w_t)$
				\ENDFOR
				\STATE $w_{t+1} \gets \sum_{k=1}^{K}\frac{n_k}{N} w_{t+1}^k$
				
			\ENDFOR	
		\end{algorithmic} 
	\label{alg:FedAvg}		
	\end{algorithm}

	\begin{algorithm}[H]
	\caption{$ClientUpdate(k, w)$ as in~\cite{mcmahan2017communication}. \newline $B$ is the local minibatch size, $E$ is the number
		of local epochs, and $\eta_c$ is the learning rate.}
	
	\begin{algorithmic}
		\STATE \COMMENT{At client $k$}
		\STATE $\mathcal{B} \gets (\mbox{split $P_k$ into batches of size $B$})$ 
		\FOR {each local epoch $i$ from $1$ to $E$}
			\FOR {batch $b \in \mathcal{B}$}
				\STATE $w \gets w - \eta_c \nabla l(w;b)$
			\ENDFOR
		\ENDFOR
		\RETURN  $w$ to server
	\end{algorithmic} 
	\label{alg:ClientUpdate}
\end{algorithm}
	
	In some variants, instead of the updated weights, $w_{t+1}^k$, each client would send the difference, $w_{t+1}^k - w_t$. This quantity is nothing but the summation of the batch gradients for each client multiplied by the negative of step size
	\begin{equation}
		\label{eqn:weight_diff}
		w_{t+1}^k -w_t = -\eta_c \sum_b \nabla l^k(w;b)
	\end{equation}
	where $\nabla l^k(w;b)$ denotes the gradient of client $k$ with batch $b$. The aggregator would calculate $f(gradient) = 
	\sum_k \frac{n_k}{N}	w_{t+1}^k -w_t$ and use equation~\eqref{eqn:weight_update} to update the global weights. When $\eta_c = 1$, we effectively see that the update is equivalent to $w_{t+1} \gets \sum_{k=1}^{K}\frac{n_k}{N} w_{t+1}^k$ as in \texttt{FederatedAveraging}. The advantage of looking at the optimization as presented in ~\eqref{eqn:weight_update} and client updates as in~\eqref{eqn:weight_diff} is that we could employ any of the gradient optimizations like Adam, SGD, RMSProp~\cite{sun2019survey} both at client and server, which gives more handle to tuning weights in non-IID settings.
	
	We now see how the distribution of the federated data plays a role in shaping the federated ``learning''.
	\subsection{Federated Data}
	In the federated setting, the "data" is held at individual end devices. Depending on the learning problem, this data could be large or small in number, multi-labeled or single labeled  and could even be restricted to a single sample. We look at each of these cases seperately in this section.
	
	\subsubsection*{Non-IID data}
	When we have a large number of samples, like in the case of MNIST~\cite{cohen2017emnist}, or next word prediction~\cite{hard2018federated}, we cannot assume the samples present in devices to be IID, as most of the time, the data distribution would be dependent on the user of the device. 
	In the case of centralized models, the mini-batch data is randomly selected, and hence we could assume $E[\nabla f_B(w)] = \nabla f(w)$. We saw that when $C=1$, \texttt{FederatedSGD} actually solves the full training data optimization. When $C<1$, although the cost function is similar to SGD, the non-IID nature of the data makes the learning skewed to the characteristics of the participating devices. However, it is shown~\cite{bonawitz2019towards, nilsson2018performance} that with more number of aggregation rounds, it will converge, but may be slightly off from the optimum obtained in central training.
	
	In \texttt{FederatedAveraging}, even when $C=1$, if the number of epochs $E$, or the number of batches is more than one in $ClientUpdate$, the expected loss becomes different from the full training loss because of non-IID data. The optimization solution is also different as the gradients calculated by end devices in $ClientUpdate$~\eqref{eqn:Fed_wt_update} are over batches and epochs, locally. Thus the gradient update from each client is very specific to the non-IID data held by the devices. If the end device data distribution is similar to the global distribution, then  learning would be similar to centralized SGD steps. If the end device data distribution is quite dissimilar to the global distribution, then the gradients can be quite noisy to each other, which may slow up and hinder the learning.

	We consider two extreme cases where the local data distribution is very different from global distribution: (1) multi samples, single label (2) single sample (single label). These types of data are common with user profile predictions  like demographics, interests, etc.
	
	\subsubsection{Multi sample single label}
	For device centric profiles that does not change over time, the profile labels would be unique to a device. This could be either demographics attributes which change very rarely or user interest profiles, which may drift very slowly. For both the cases, we could consider that the label is unique (atleast for a predetermined time period). The features however may be thought of as a single feature vector or a group of vectors denoting, for instance, the hour, day, week or month.
	In such cases, we have multiple samples in each device all with the same label. Thus the updated weights from each device $k$, $w_{t+1}^k$ would be directed more towards making the activation corresponding to its own label. In other words, devices with unique labels learn weight updates very specific to their attached labels. The server aggregation would consist of aggregating many "label overfitted" weight updates. We will show this effect more clearly in our experiments section.
	
	\subsubsection{Single sample (single label)}
	This is an extreme data distribution scenario where there is only one sample and hence one label in each device. However for the
	case, when $E=1$, and with random selection of end-devices, this would actually correspond to mini-batch SGD! The main problem here is that, in federated learning, communication is costly (mainly because of the time, and also the payload), and local computation is cheap. Hence federated setups try to synchronize as many devices as possible in each aggregation. It is shown that $400\sim1000$ devices per round works well both in terms of cost and in terms of learning~\cite{bonawitz2019towards}. In single sample scenario, the number of devices would correspond to the  batch size in SGD. The batch size is a hyper parameter in SGD, which depends on the problem in hand.  It has been observed in practice that when using a larger batch there is a significant degradation in the quality of the model, as measured by its ability to generalize~\cite{keskar2016large}. The general batch sizes are $32\sim 250$.

	\subsection{Effects of model complexity on learning and cost}
	\begin{table*}
		\small
		\centering
		\caption{Models in rows 1-3 are found in the literature. Models in rows 4,5 were designed by us taking into account the size factor. Model2 has 3 convolution layers and 2 fully connected layers. Model1 has a Global MaxPool layer (denoted as GMPool) in between the convolution and fully connected layers which drastically reduces the model size. Sixth row is Model2 trained in an FL way involving 10 devices.}
		\label{tab:image_image}
		\begin{tabular}{@{}lcccc@{}}\toprule
				\# & Dataset & Method & Accuracy & Size(MB) \\
				\midrule
				1 & ImageNet(pretrain)+IMDB-WIKI( fine tuning) Adience(testing) & 	AL-ResNets-34~\cite{zhang2019fine} &  66.03&  ~87.3\\
				2 & ImageNet(pretrain)+Adience(training and testing) & RESF-EMD~\cite{hou2016squared} & 62.2 & ~103 \\
				3 & ImageNet(pretrain)+IMDB-WIKI( fine tuning) Adience(testing) & VGGF-DEX with IMDB-WIKI~\cite{rothe2018deep} & 64  & ~553 \\ \midrule
				
				4 & IMDB-WIKI~\cite{rotheimdb} dataset & Model1 (4Conv+GMPool+ 2FC) & 46.83 & 1.61 \\
				5 & IMDB-WIKI dataset & Model2 (3Conv+2FC) & 49.34 & 24.03 \\
				6 & IMDB-WIKI dataset & Model2 FL & 45.23 & 24.03 \\
				
				\bottomrule
			\end{tabular}
		\end{table*}
		
	Model complexity is an important parameter for any learning task. An optimal model for any problem would decrease both bias and overfitting, and hence generalizes well. The complexity of the models is largely governed by the complexity of the features to be learned from the input. Or in other words, when we have more features inherent in the data, more complex models would suit better. In central schemes, there is not much restriction of the model complexity, in terms of the cost involved in the training. The main cost of  training is on the number of compute instances involved and the duration of training. The model complexity is not much of a factor although we could assume complex models to have lengthier training periods. 
	However there is no communication cost that is dependent on the model. 
	
	In federated setting, there is a communication cost  associated with passing model parameters and configurations up and down the server to clients. The more the size of the model, the more the communication cost as is shown in section~\ref{sec:cost}. In addition, devices may not have the computation power left aside for doing computationally intensive trainings. Thus learning in federated setting would be challenging as it needs to compromise between performance and model size (cost), in many scenarios.
	
	Take the example of image classification. Table~\ref{tab:image_image} shows the performance and model sizes of various age prediction models from images. Usually, age prediction is considered as a classification problem, where the intention is to classify subjects into different age buckets. The purpose of table is not a comparison of models, but to understand the accuracy levels achieved by different models of different complexities (sizes). As we see, complex models evidently perform well for the task, however the models may not be adopted for federated learning as the communication costs would be immensely high. For instance, using the assumptions mentioned in section~\ref{sec:cost}, the training cost for low performing, low sized models, Model1 and Model2 in table~\ref{tab:image_image} would be $\$ 1111$ and $\$ 6290$ respectively. For heavy models AL-ResNets-34, RESF-EMD, VGGF-DEX, the training costs are $ \$20905, \$24532,\mbox{ and } \$128482$ respectively! In addition there would also be a huge deployment cost, that would be directly dependent on the model size.
	
	Hence, unlike central settings, it is very important to look at model sizes when we try for federated learning. More emphasis should be given to obtaining most relevant and small subet of features, as well as other model size reduction techniques like pruning or quantization~\cite{han2015deep}.
	
	
	\vspace{5pt}
	\section{Experimentation and results}
	\label{sec:exp}
	We now show the experimental results for the different scenarios we discussed in the federated learning setup. We specifically show how the number of samples and labels per device have an effect on overall learning.
	\subsection{Data set and features}
	The dataset we used is the MNIST data for handwritten digit recognition~\cite{lecun1998mnist, lecun1998gradient}.The MNIST dataset consist of 70000 handwritten images ($28\times 28 \times 1$) of the 10 digits. 60000 samples are used for training and 10000 samples  for testing. The images were flattened to a 784 dimension vector. The hardware setup was made up of a Intel(R) Core(TM) $i7-4770K$ CPU \@ $3.50GHz$ processor with 32 GB RAM running on a  Ubuntu 16.04.7 LTS operating system. No GPUs were involved in the simulations. The central models were trained using Keras~\cite{chollet2015keras}, and the federated training was simulated using Tensorflow Federated~\cite{The_TensorFlow_Federated_Authors_TensorFlow_Federated_2019}.  
	`
%

	\subsection{Number of samples per device}
	\begin{figure}[h!]
		\centering
		\includegraphics[width=0.75\linewidth]{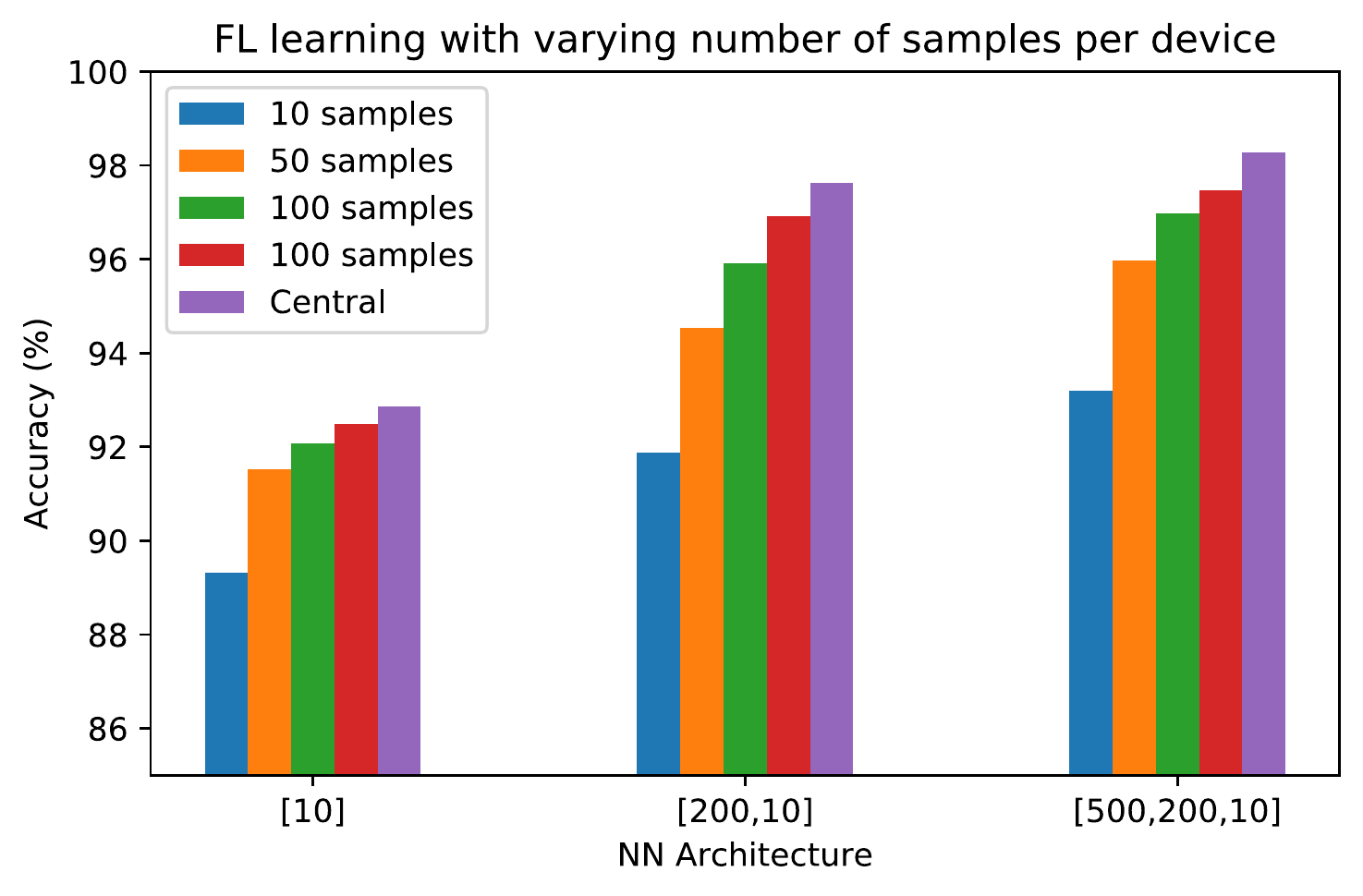}
		\caption{Change in accuracy with number of training samples per device. We see that as the number of samples increases, the test accuracy increases. Also as the number of neural network layers are increased, the federated learning training also shows an improved learning performance.}
		\label{fig:numsamples}
	\end{figure}
	In federated learning, the number of training samples per device have a direct effect on training. Fig.~\ref{fig:numsamples} shows the change in accuracy with varying number of training samples per device. We train three models with (a) no hidden layer, (b) single layer with 200 nodes, (c) two hidden layers with 500 and 200 nodes. 
	We represent these networks as $[10], [200,10], [500,200,10]$ respectively 
	\footnote{[500,200,10] represents a neural network with three layers. Two hidden layers with 500 and 200 nodes each and output layer with 10 nodes.}. For comparison, we also trained these models centrally.
	
	As is the case with centralized schemes, the more the number of samples per device, the more generalized the per device weight updates would be. With fewer samples per device,  the gradient updates coming from individual devices can contain more noise. And hence there would be greater variance in the centralized updated weights. When we increase the number of layers, the FL learning is consistent with the centralized models in learning better.
	
%
%

	\subsection{Learning with single label}
\begin{figure}
		\centering
		\includegraphics[width=0.75\linewidth]{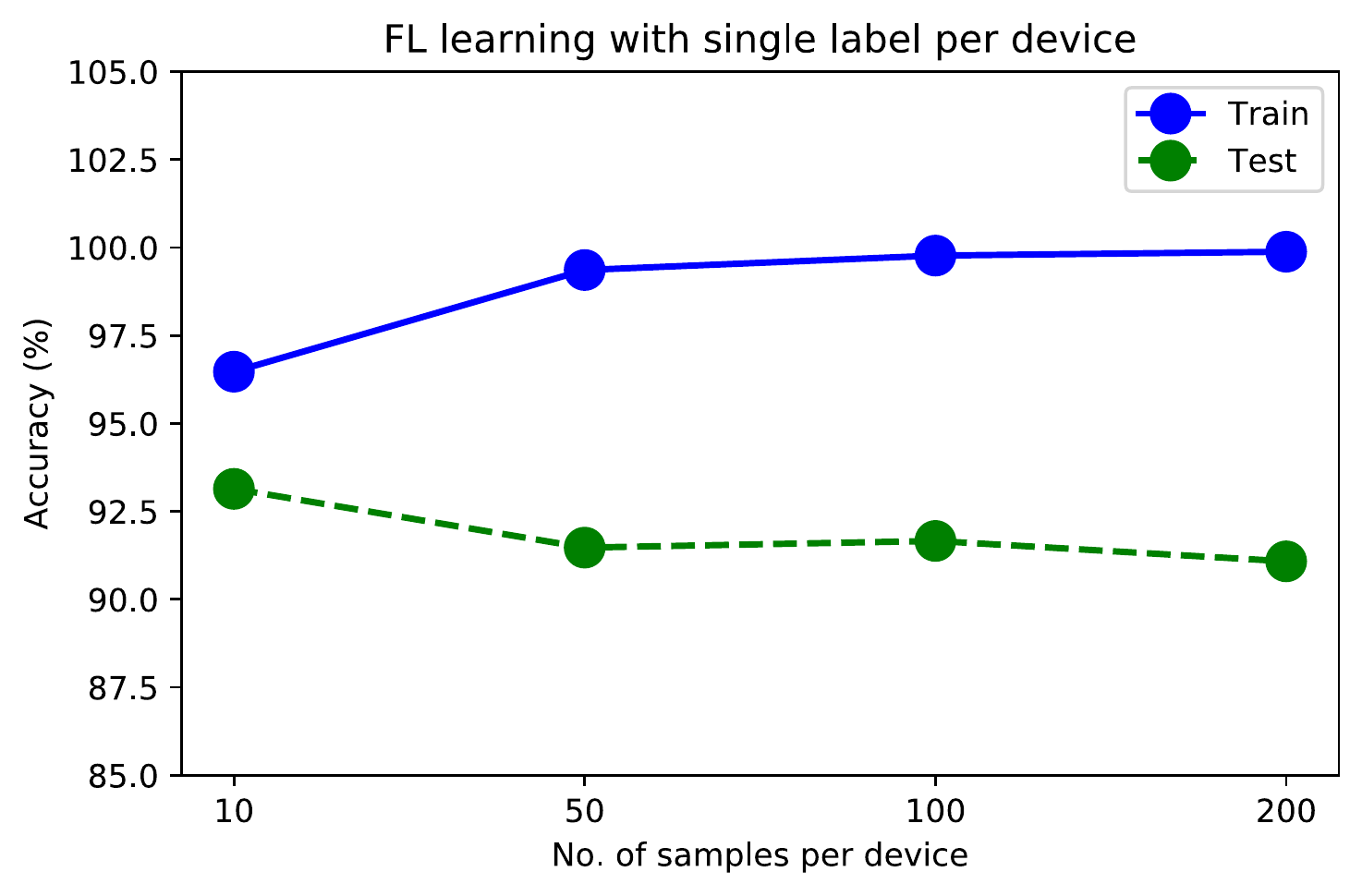}
		\caption{Single Label: Training with  many samples per device with single label on the [500,200,10] network. We see that as the number of samples increases, the train accuracy is high, but the test accuracy does not improve. This shows overfitting of samples.}
		\label{fig:Onelabel}
	\end{figure}

	When there are multiple samples per device, all with the same label (but different for different devices), we see that the learning takes an opposite course, i.e., better learning is obtained with fewer number of samples. Fig.~\ref{fig:Onelabel} shows the training and test accuracies for the [500,200,10] network architecture with $10,50,100\mbox{ and }200$ samples, all with the same label per device. 
	We see that for 10 samples per device, the training and testing accuracies are comparable. With more number of samples, we see a scenario of clear overfitting. So how does this overfitting occur, which was not observed when we had all classes in each device? 
	
\begin{figure}
		\centering
		\includegraphics[width=0.75\linewidth]{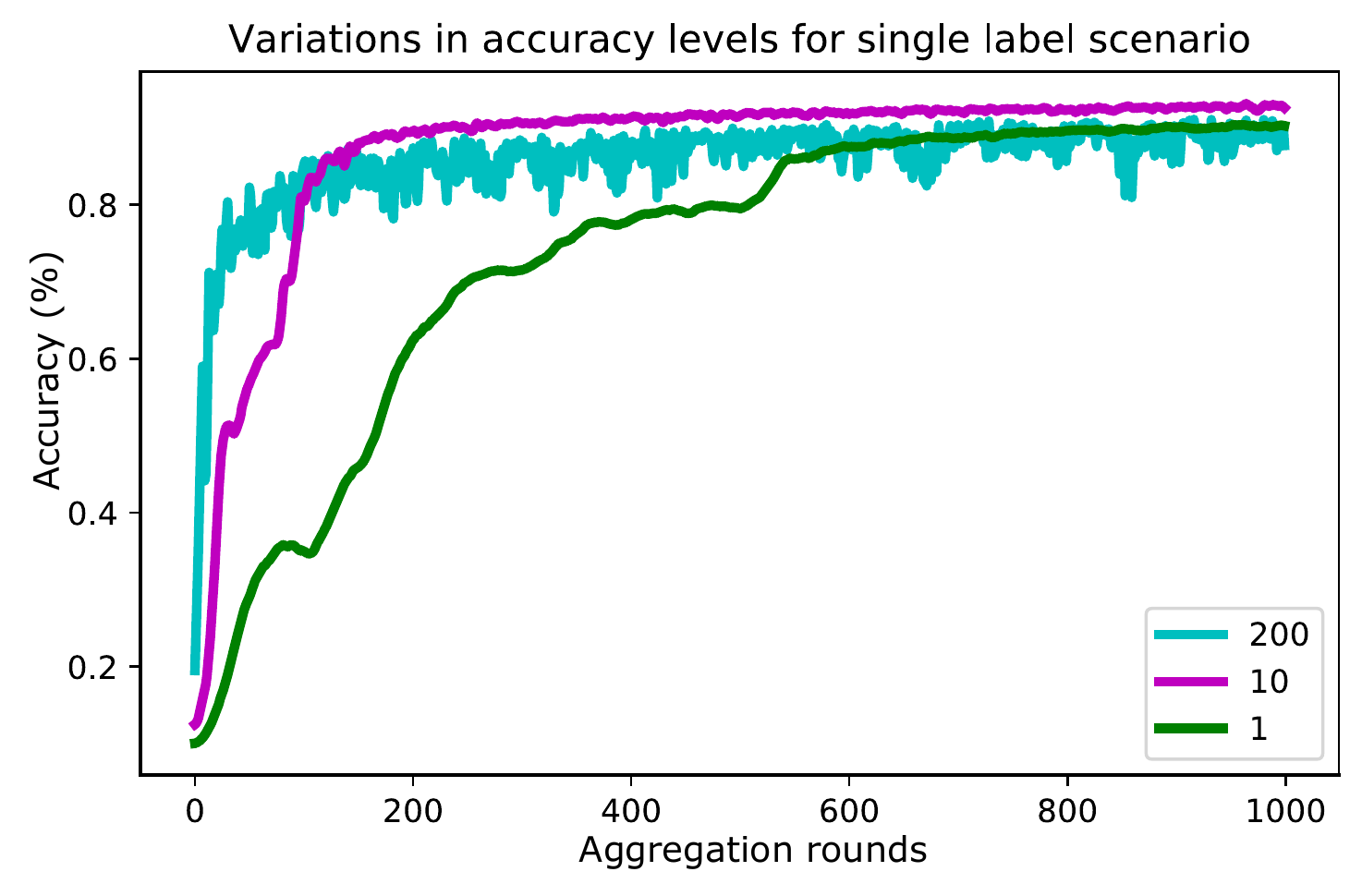}
		\caption{Variation in accuracies: The validation accuracies over the aggregation rounds, when we have multiple samples all with the same label for each device. We compare the learnings over rounds for 10 samples per device and 200 samples per device. When the number of samples are less, then the label specific learning is less and hence there is more generalization. With more samples with same label per device, there is overfitting of labels within device. This leads to noise on aggregation. For single sample, we see a scenario of underfitting.}
		\label{fig:OnelabelAccuracies}
	\end{figure}

	Note that each device learning is specific to the single label attached to the device. Hence each device weights would be tuned (overfitted) to match only that label available on it. Thus the server would be aggregating different ``label overfitted'' gradient updates. This characteristic is more evident in Fig.~\ref{fig:OnelabelAccuracies}. When the number of samples are ten, there is less label specific learning, and hence better generalization, in comparison to when we have more samples (200) per device. 
	\subsubsection*{Single sample (single label)}
		In the single label case, while we see overfitting with increasing number of samples per device, we see underfitting, when the number of samples is too low or one. This is evident in Fig.~\ref{fig:OnelabelAccuracies}, where the learning curve is slow for single sample cases as compared to the two other multi-sample single label cases. Note that as with ten samples, the variation in accuracies for single sample over rounds is also relatively less as compared to more number of samples.

	\section{Cost assessment}
	\label{sec:cost}
	In this section, we compare the costs for training a centralized model against federated learning settings. We consider both training and deployment cost for the respective approaches.
	For the cost comparison between federated learning and centralized schemes, we use the actual Amazon Web Services (AWS) cost figures, at the time of
	publishing of this work~\cite{aws_pricing}. 
	
	\subsection{Federated Learning communication}
	\begin{figure*}[ht]
		\centering
		\includegraphics[width=\linewidth]{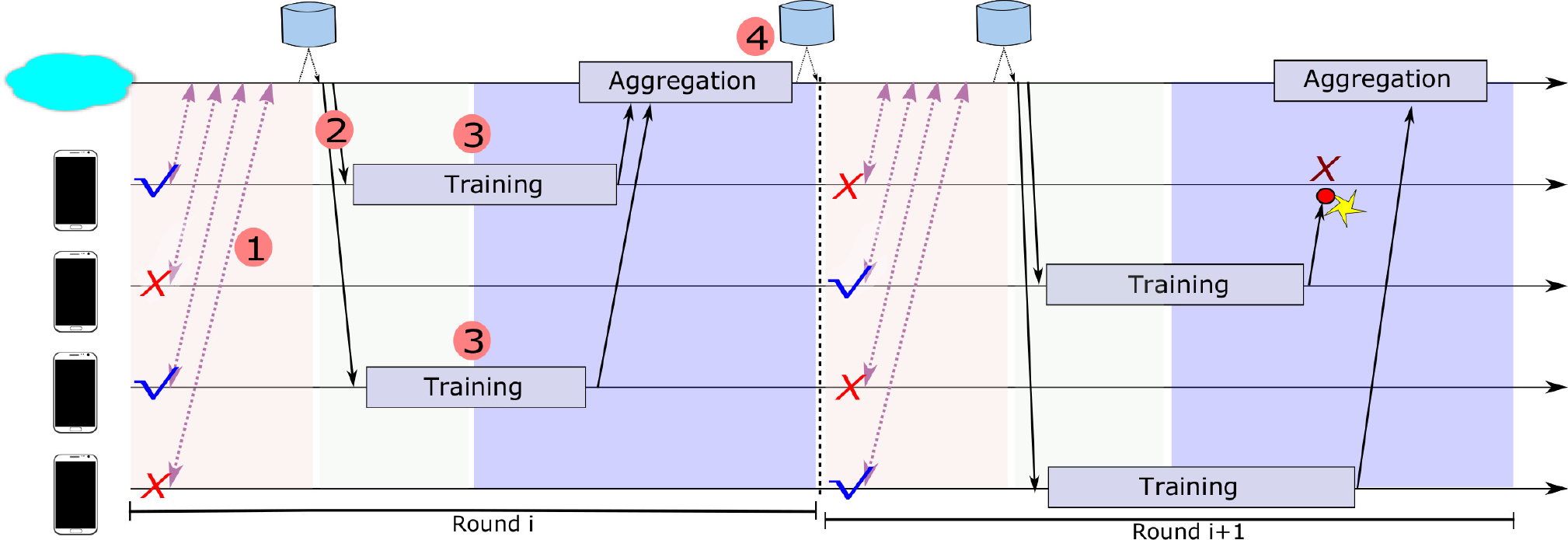}
		\caption{A typical federated learning communication sequence.}
		\label{fig:FLproto}
	\end{figure*}
	For federated learning costs, we assume the communication mechanism presented in~\cite{bonawitz2019towards, li2020federated}. For calculating the cost, we consider the following steps for each aggregation round:
	\begin{enumerate}
		\item
		\begin{enumerate}
			\item The registered devices which are ready, ping the server to participate in a training cycle.
			\item The server selects a subset of these devices. 
		\end{enumerate} \label{item1}
		\item The server sends the  current model with weights and training plan, which could be thought of as a executable that the device needs to run on its data on the model.
		\item The devices undergoes local weight updation and the updated weights are send back to the server.
		\item The server aggregates these weights and updates the global model. \label{item2}
		\item[*] The above steps are repeated until a server decided convergence criteria.
	\end{enumerate}
	The above steps~\ref{item1}-\ref{item2} are shown in Fig.~\ref{fig:FLproto}. Step~\ref{item1}, could be either initiated by the client or server. The most cost effective scenario would be that the server pings a subset of registered devices for its readiness to participate in each training cycle. This would however require additional logic on the side of server to choose the most plausible devices that could participate in each round. 
	
	In centralized setup, communication and data transfer is required for data syncing to the S3. The computation is done via EC2 instances, and the inferred labels are synced back to the devices. In the case of FL, the communication and data transfer is mainly for transfer of model weights and updates from/to server to/from devices.  
	
	For the assessment, we are assuming, there are around 500K devices, that is registered for FL training, and the total population is 12 million. The number of devices participating in each of round of training is fixed at 500. The training is assumed to complete within a month and the cost computation is calculated for a month.
	
	We assume the ping messages and other request messages to be atmost 1KB. The main variables in a FL setup are the model size, number of rounds for training and number of rounds of training per day (inversely proportional to the time per round). For the FL training, the communication and model and weight download/upload is assumed to happen between devices and EC2 (server aggregator). For the final model deployment, the devices download the model from S3 directly, as it found to be more efficient than getting pulled from EC2.  We account the data transfer payload cost from AWS to internet and the S3 storage, read\footnote{S3 storage and write are minimal, while the main cost would be for read corresponding to the model reads during the deployment} and write. For the EC2, we account for Portal(m4.xlarge, 1)\footnote{The number specifies the number of instances.} Model aggregator	(c5.xlarge), Training Server	(m4.xlarge), Monitoring Node (t3.xlarge) and Jump Box (t3.medium). The model aggregator and training server are accounted only for the duration of the training and rest for the entire month. An ELB (accounted during training), Route53 components and VPC (NAT gateway) are also considered.
	
	We assume a similar infrastructure for centralized setting except for EC2. For the EC2, we account for Portal (m4.xlarge, 1)
	Ingestion machines	(m4.xlarge, 3), Training Server (m4.xlarge, 7), Tagging server	(m4.xlarge, 3), Monitoring Node	(t3.xlarge, 1), 
	Jump Box (t3.medium, 1). For the training, we assume 6 instances are employed for a day for data engineering (feature generation) and 2 instances are employed for a day for training. The tagging is assumed for 4 days. The main cost factor for centralized training is the data syncing cost. We are assuming that the 12 million devices sync on an average 250 KB of data monthly per device. The label size to be synced back is assumed to be 100 Bytes per device. With these figures, the centralized training  cost is $\$ 1967.90$. To understand the effect of data sync, when the average data that is synced is assumed to be 1 MB (data that contains logs specific to training), which is a reasonable value, the training cost is $\$3769.70$. There is no "deployment" as  such, as the centrally inferred labels are synced back to the devices, which is found to be negligibly small (Less than \$1 in our computations).

	In the analysis, we have not accounted for communication losses. Also the ELB and EC2 instances for FL may not be optimized and can be done so based on the load requirements.
	
	\subsubsection*{Factors affecting cost}
	The main factors that affect the FL cost are the message payloads, the weight transfer payloads and training time. The training time determines the time for which the training related EC2 instances have to be kept live. To simplify the analysis, these three factors are captured using three metrics (a) Model size (b) Number of aggregation rounds (c) Number of aggregation rounds per day.
	The model size affects data payload, both during training and deployment, number of aggregation rounds affects messages and data payloads, and the training time, the number of aggregation rounds per day affects the time for keeping the training EC2 instances live.
	
	
	\begin{figure}
		\centering
		\includegraphics[width=0.75\linewidth]{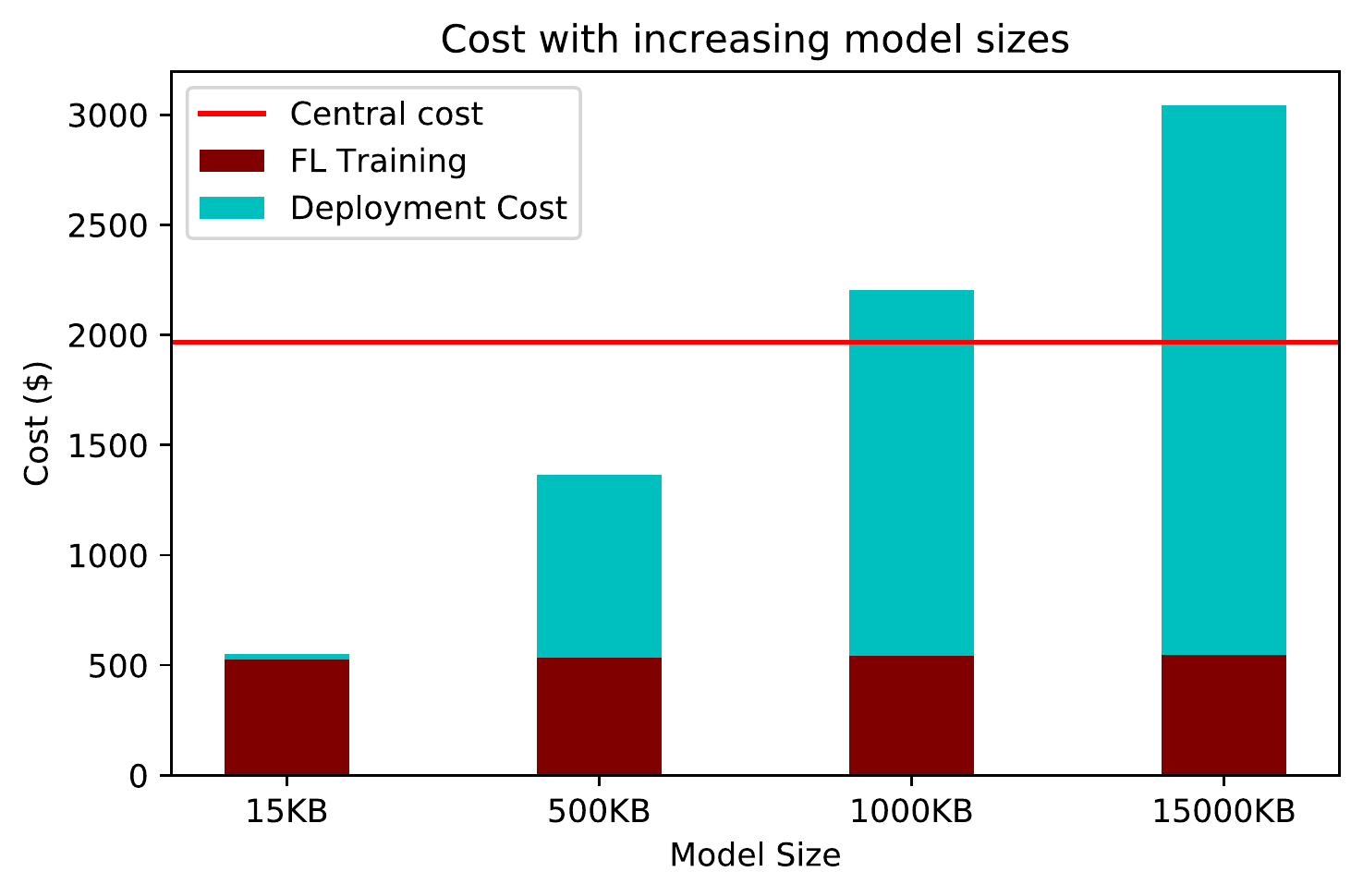}
		\caption{Cost as a function of model sizes. The assumed number of rounds is 200 and the number of rounds per day is 100.}
		\label{fig:modelsize}
	\end{figure}
	Fig.~\ref{fig:modelsize} shows the effect of model size on cost. The training cost is almost same in all the cases (525.68, 533.15, 540.85, 48.55 for model sizes `15KB', `500KB', `1000KB', `15000KB' respectively). However the deployment cost takes a huge hit when the model sizes are increased. This is because of the S3 reads that need to be done for all the devices to which the model needs to be deployed. Hence it would be advisable to look for alternative ways of deployment, that does not involve S3 reads, like a FOTA deployment or a release deployment.
	
	The impact of model size on cost also leads to thinking of model size reduction techniques. There are two ways in which model size could be reduced: (1) careful model design including feature set reduction (2) model compression techniques. Many a times, the size of a feature set could have an impact on the model size. For instance, if we are considering bag of words or similar techniques with large vocabulary, then this would increase the size of the model. Dimensionality reduction techniques like Principal Component Analysis~\cite{wold1987principal} may not be applicable in federated setting, as the entire data is not available for the matrix analysis. Techniques like hashing~\cite{weinberger2009feature} could be utilized for this purpose and the performances are found to be close to the non-hashed features performance. There are also techniques~\cite{zhang2018lq, cheng2020survey, han2015learning} for keeping neural networks compact which not only keeps the models small, but make the learning fast.~\cite{cheng2020survey} discusses the various techniques in this front. ~\cite{han2015learning} proposed a well adopted pruning technique, in which the neural network evolves over time to converge to the best compact neural network.
	
	\begin{figure}
		\centering
		\includegraphics[width=0.75\linewidth]{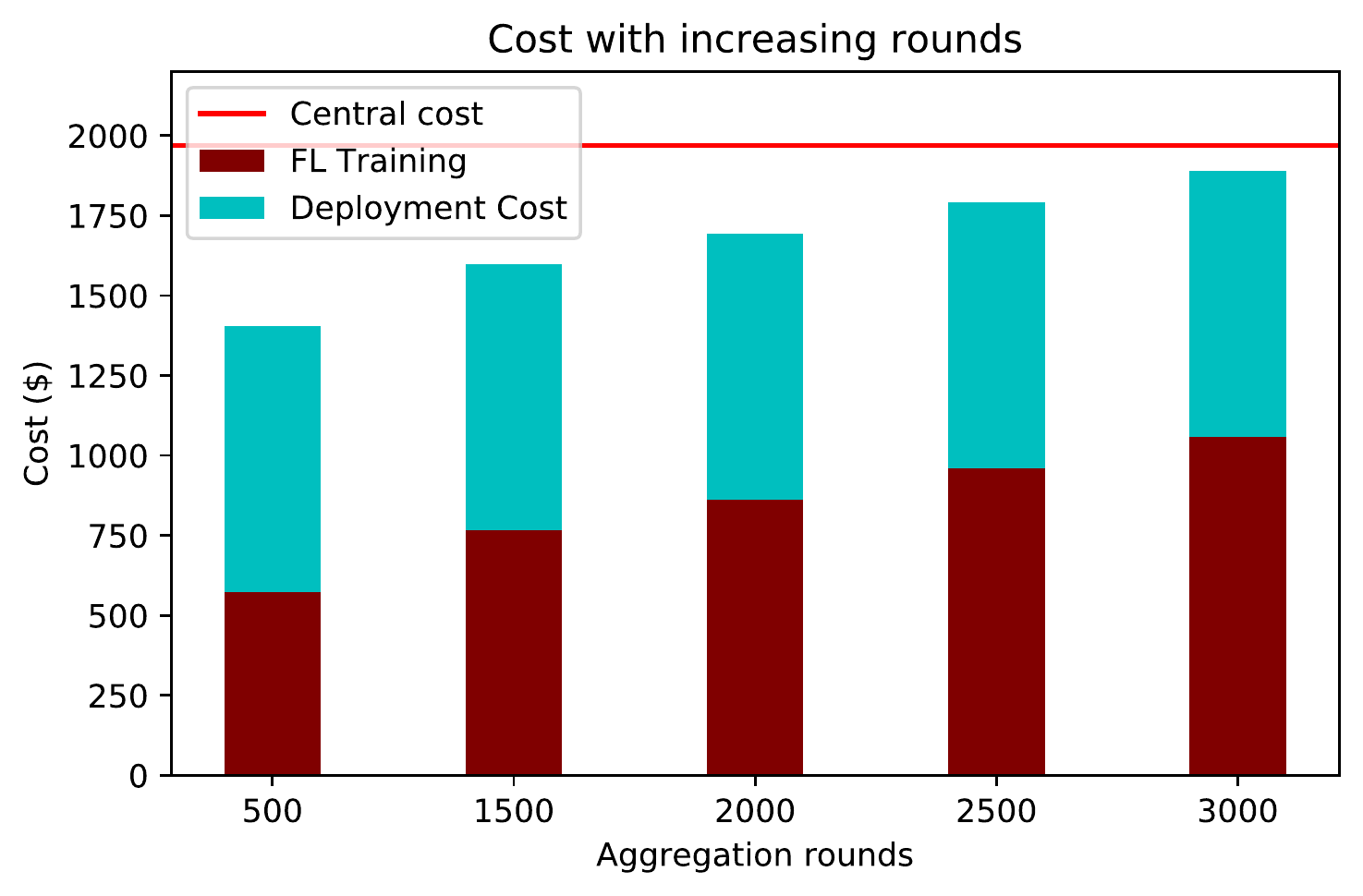}
		\caption{Cost as a function of number of aggregation rounds. The assumed Model Size is 500 KB and the number of rounds per day is 200.}
		\label{fig:numrounds}
	\end{figure}
	In Fig.~\ref{fig:numrounds}, we show the effect of the number of rounds on cost, specifically on the training. The number of rounds would have an impact on the training duration as well as the total number of communication messages that are passed between clients and servers, including the model weights and updates.
	
	\begin{figure}
		\centering
		\includegraphics[width=0.75\linewidth]{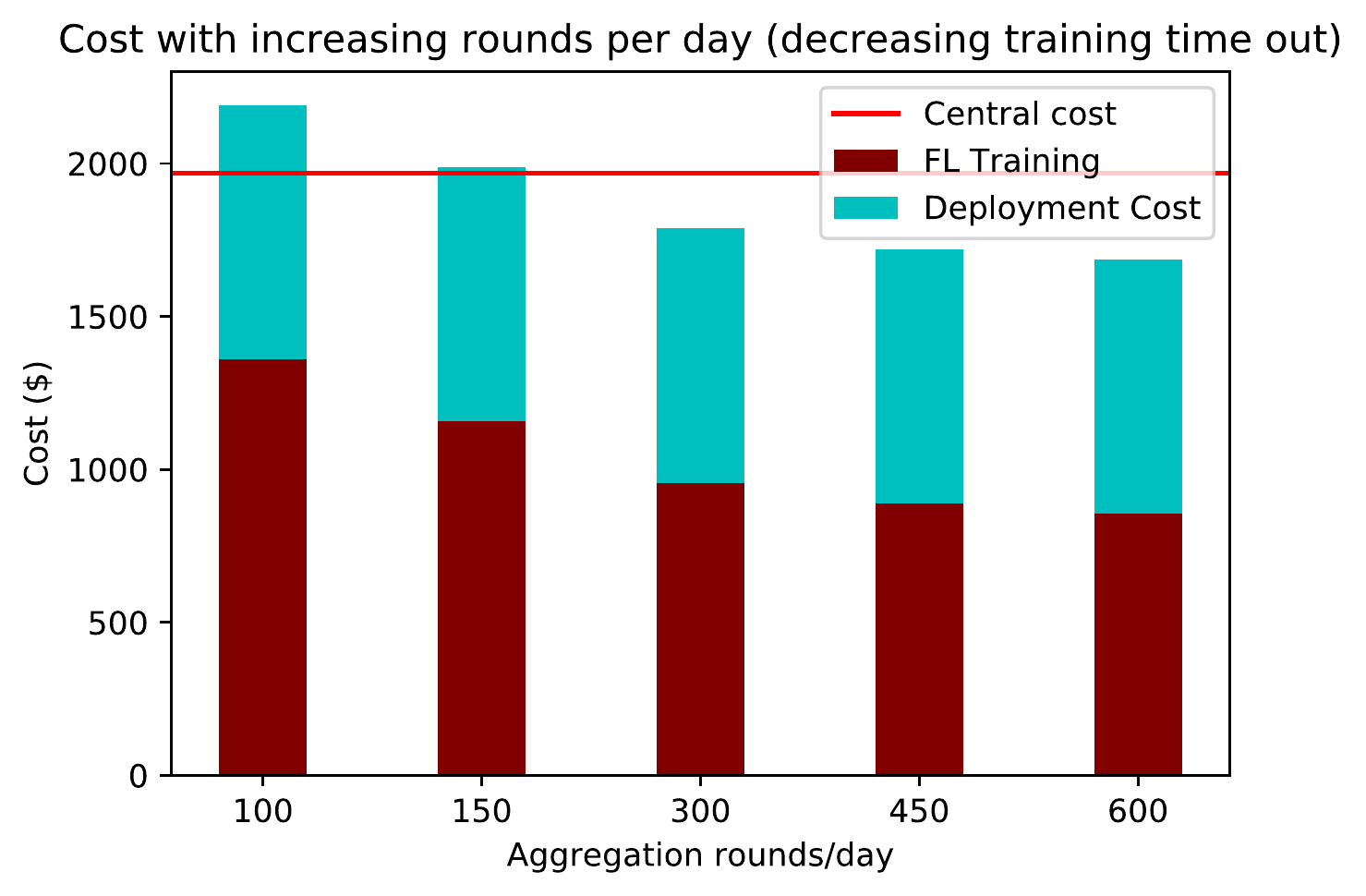}
		\caption{Cost as a function of number of aggregation rounds per day.. The assumed model size is 500 KB  and the number of rounds  is 3000.}
		\label{fig:roundsperday}
	\end{figure}
	The effect of the number of aggregation rounds per day is shown in Fig.~\ref{fig:roundsperday}. The cost decrease is because of the reduction in EC2 training instances live period. The more the number of aggregations per day, the less the number of days we need to keep the training instances running. This factor is affected by the per cycle timeout that can be specified by the FL engineer during training. Other factors could also play a role in restricting this number, like shortage of participating devices during certain periods of the day (like work time), the model complexity, which makes each gradient iteration at clients slower, etc.

	
	\vspace{3pt}
	
	\section{Conclusion}
	In this work, we have looked more closely into federated learning training for non-IID cases, where there is large skewness in labels. We showed a comparison between the different SGD schemes with federated SGD and federated averaging in different data distribution scenarios. We showed that single label per device scenario does not work similar to multi label cases, and care should be taken about the optimal samples used for training. Too many samples with same labels may actually hamper performance. We also showed the cost comparison of FL with centralized schemes and looked at factors that affect cost, viz, model size, number of rounds and number of rounds per day. We showed that in FL, model deployment has a huge affect on cost and product owners need to look how to efficiently manage deployment.
	\section{Acknowledgement}
	We would like to thank the Samsung R$\&$D institute for providing the resources for performing this work. 
	
%

	\bibliographystyle{unsrt}
	\bibliography{refs}

\end{document}